\documentclass[letterpaper, 10pt, conference]{IEEEtran}

\usepackage{amsmath}
\usepackage{amssymb}
\usepackage[pdftex]{graphicx}
\graphicspath{{figures/}}
\usepackage{hyperref}
\hypersetup{
    colorlinks=true,
    linkcolor=black,
    citecolor=black,
    filecolor=black,
    urlcolor=black,
}
\usepackage[T1]{fontenc}
\usepackage[utf8]{inputenc}
\usepackage{csquotes}
\usepackage[english]{babel}
\usepackage[skip=1mm,font=small]{caption}
\usepackage{subcaption}
\usepackage{placeins}
\usepackage{pgfplots}
\usepackage{multirow}
\usepackage{pgfplotstable}
\usepackage{booktabs}
\usepackage{multirow}
\usepackage{array}
\usepackage{siunitx}

\newcommand{\norm}[1]{\left\lVert#1\right\rVert}

\pgfplotsset{
compat=1.12,
width=0.98\columnwidth,
height=4cm,
xmin=0,
ymin=0,
cycle list name=exotic,
log plot exponent style/.style={/pgf/number format/precision=0},
}
\usepgfplotslibrary{groupplots}
\newcommand{\readcsv}{\pgfplotstableread[col sep=comma, ignore chars={"}]}

\title{
Capturing Object Detection Uncertainty in Multi-Layer Grid Maps
}

\author{\IEEEauthorblockN{Sascha Wirges \& Marcel Reith-Braun}
\IEEEauthorblockA{Mobile Perception Systems Group\\
FZI Research Center for Information Technology\\
Karlsruhe, Germany\\
wirges@fzi.de}
\and
\IEEEauthorblockN{Martin Lauer \& Christoph Stiller}
\IEEEauthorblockA{Institute of Measurement and Control Systems\\
Karlsruhe Institute of Technology (KIT)\\
Karlsruhe, Germany\\
\{lauer,stiller\}@kit.edu}}

\begin{document}

\maketitle
\thispagestyle{empty}
\pagestyle{empty}

\begin{abstract}
We propose a deep convolutional object detector for automated driving applications that also estimates classification, pose and shape uncertainty of each detected object.
The input consists of a multi-layer grid map which is well-suited for sensor fusion, free-space estimation and machine learning.
Based on the estimated pose and shape uncertainty we approximate object hulls with bounded collision probability which we find helpful for subsequent trajectory planning tasks.
We train our models based on the KITTI object detection data set.
In a quantitative and qualitative evaluation some models show a similar performance and superior robustness compared to previously developed object detectors.
However, our evaluation also points to undesired data set properties which should be addressed when training data-driven models or creating new data sets.
\end{abstract}

\section{Introduction}
\label{sec:introduction}

As part of scene understanding a key component of safe and robust mobile robotic systems is the identification of relevant objects which the system may interact with.
In the context of automated vehicles scene understanding may include the detection, shape and motion estimation as well as semantic classification of relevant traffic participants such as pedestrians, cyclists and cars.
Although tremendous progress in this field was made in recent years, scene understanding is still far from being perfect.
Reports \cite{DMV2017} on test drives with automated vehicles show that there is still a considerable amount of manual driving disengagements by human test drivers instead of the automated system.
Therefore it is important to make a system capture its uncertainty in order to detect false estimates.
This way, the system is able to transit automatically into a fail-safe state or hand-over to a human driver.

Here, we focus on uncertainty estimation in object detection, pose- and shape estimation and classification, referred to as \textit{object detection}.
We represent the scene by a multi-layer top-view grid map which provides a mapping from two-dimensional discretized ground surface coordinates to higher-dimensional features.
First introduced in \cite{Elfes1989}, grid maps are well-suited for sensor fusion and enable the use of efficient convolutional operations due to their dense grid structure.
Using range sensors (e.g. lidar), features might be composed of the number of surface reflections, minimum and maximum height above ground or average reflection intensity.
Geometric information such as the number of free-space observations per cell or the height of cast shadows can be incorporated by casting rays from the sensor origin to the reflection positions.

\begin{figure}[t]
\includegraphics[width=\linewidth]{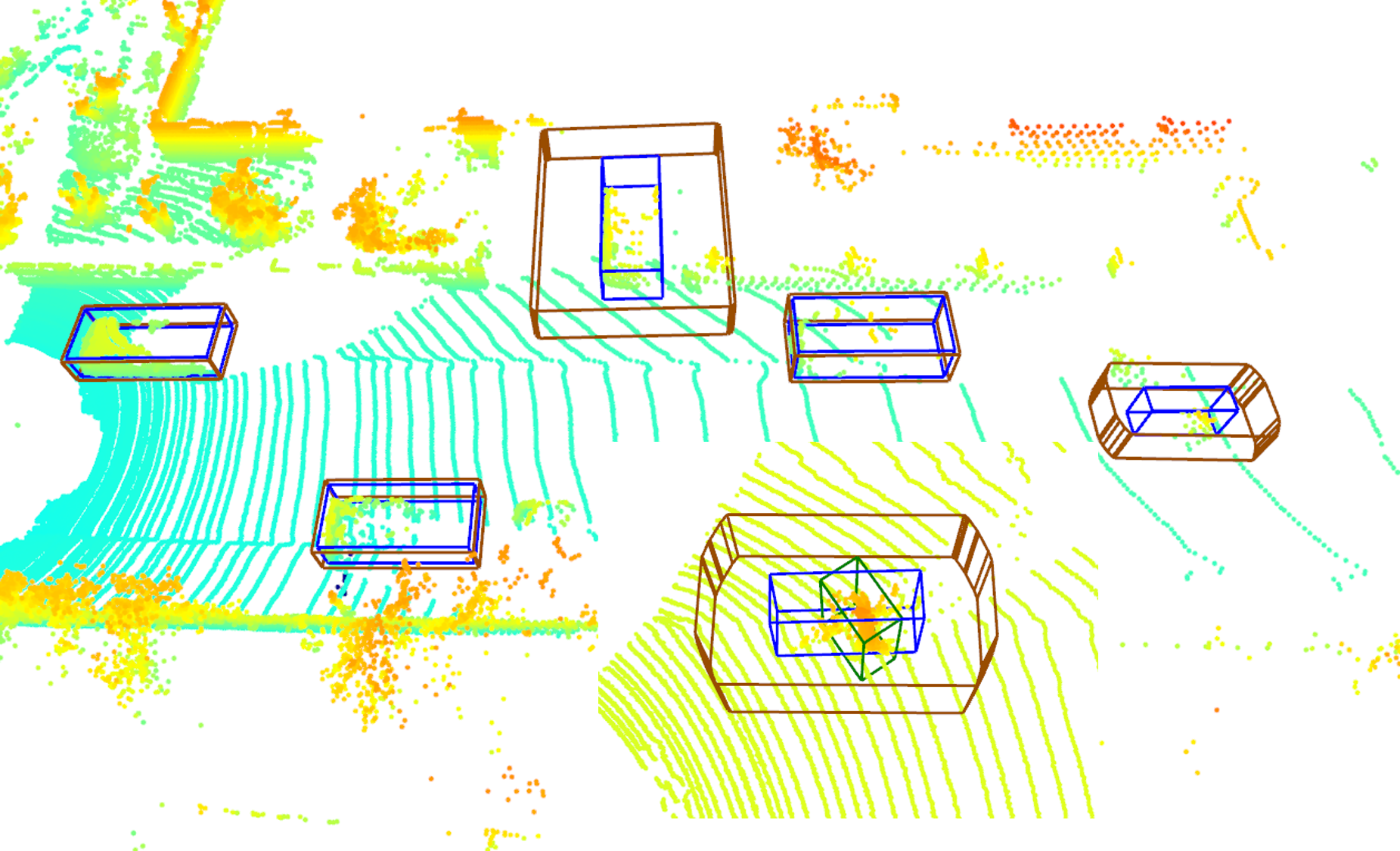}
\caption{
Top view of detected objects and their shapes resulting from uncertainty estimates.
Green: Ground-truth, blue: median bounding box, red: convex hull for 95th percentile.
We observe a high epistemic localization uncertainty for pedestrians and cyclists, resulting in larger shapes.
On the other hand, cars orthogonal to the driving direction yield large uncertainty due to their underrepresentation in the data set.
}
\label{fig:top_view}
\end{figure}

Given the scene representation as grid map, our system outputs a list of relevant objects, each of it assigned with class, pose, shape estimates and uncertainties, respectively.
Therefore, we start by introducing uncertainties in neural networks and recent estimation techniques in Section~\ref{sec:related_work_modeling_uncertainty_in_neural_networks} and discuss work on capturing uncertainties in deep convolutional object detectors in Section~\ref{sec:related_work_uncertainties_in_object_detection}.
After introducing the KITTI Bird's Eye View benchmark and related contributions in Section~\ref{sec:related_work_kitti_bev_benchmark}, we provide information on the grid maps used as model input in Section~\ref{sec:model_inputs}.
To make uncertainties available to subsequent trajectory planning we develop a simple shape representation that incorporates pose and shape uncertainty (Section~\ref{sec:model_outputs}).
Another part of this work is the qualitative and quantitative comparison of different models.
Therefore, after introducing different model configurations and the training process in Section~\ref{sec:model_network_structure_and_training_process} we highlight qualitative observations as well as quantitative results for these configurations in Section~\ref{sec:evaluation}.
Some models perform similar on the benchmark compared to our previously developed deterministic method but are capable of additionally estimating uncertainties.
Finally, we conclude our work and propose ideas for future research in Section~\ref{sec:conclusion}.

\section{Related Work}
\label{sec:related_work}

\subsection{Modeling Uncertainty in Neural Networks}
\label{sec:related_work_modeling_uncertainty_in_neural_networks}

\begin{figure}[ht]
\centering
\def\svgwidth{\columnwidth}
\begin{small}
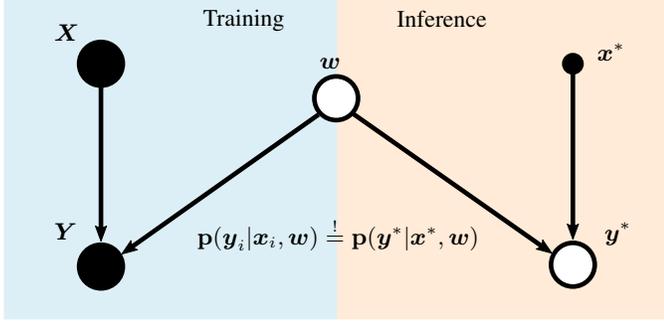
\end{small}
\caption{
Probabilistic graphical model representation of a BNN for training and inference.
During training, realizations of $\boldsymbol{X}$ and $\boldsymbol{Y}$ are used to estimate the model weights $\boldsymbol{w}$.
During inference, we aim to estimate the distribution $\boldsymbol{y}^*$ given a measurement $\boldsymbol{x}^*$ and model weights $\boldsymbol{w}$.
Therefore we assume equality between the conditioned probabilities $\mathrm{p} \left( \boldsymbol{y}_i | \boldsymbol{x}_i, \boldsymbol{w} \right)$ and $\mathrm{p} \left( \boldsymbol{y}^* | \boldsymbol{x}^*, \boldsymbol{w} \right)$.
}
\label{fig:pgm}
\end{figure}

Whereas ordinary neural networks can be interpreted as point estimators with deterministic weights, bayesian neural networks (BNNs) place distributions over their weights.
Hence, they model probabilistic relationships between inputs $\boldsymbol{x}$, weights $\boldsymbol{w}$ and outputs $\boldsymbol{y}$.
BNNs can be represented as probabilistic graphical model (see Fig.~\ref{fig:pgm}).

In the following we assume the model weights $\boldsymbol{w}$ to be estimated based on inputs $\boldsymbol{X} = \lbrace \boldsymbol{x}_i \rbrace$ and outputs $\boldsymbol{Y} = \lbrace \boldsymbol{y}_i \rbrace$ for which $N$ realizations are provided by the training data set.
Given an input $\boldsymbol{x}^*$ during inference we get the output distribution
\begin{equation}
\mathrm{p}(\boldsymbol{y}^* | \boldsymbol{x}^*, \boldsymbol{X}, \boldsymbol{Y}) = \int \mathrm{p}(\boldsymbol{y}^* | \boldsymbol{x}^*, \boldsymbol{w}) ~ \mathrm{p}(\boldsymbol{w} | \boldsymbol{X}, \boldsymbol{Y}) ~ \mathrm{d} \boldsymbol{w}
\end{equation}
by marginalizing over all weights $\boldsymbol{w}$ (see \cite{Gal2015a}).
Here, $\mathrm{p}(\boldsymbol{y}^* | \boldsymbol{x}^*, \boldsymbol{w})$ describes aleatoric and $\mathrm{p}(\boldsymbol{w} | \boldsymbol{X}, \boldsymbol{Y})$ epistemic uncertainty.
Epistemic uncertainty captures systematic errors, e.g. due to model or data set limitations.
Aleatoric uncertainty is statistical, e.g. due to noisy, occluded or distant sensor measurements.

Although marginalizing all weights is usually intractable, Gal et al. \cite{Gal2015} show that Monte-Carlo (MC) dropout can be used to approximate Bayesian inference in deep Gaussian processes.
They assume the different networks outputs $\boldsymbol{y}_i$ to be independent, identically distributed and only depending on networks weights $\boldsymbol{w}$ and input $\boldsymbol{x}_i$.
This way they can model the predicted output $\mathbf{f} \left( \boldsymbol{x}_i, \boldsymbol{w} \right)$ as function of input $\boldsymbol{x}_i$ and weights $\boldsymbol{w}$.
The authors assume the output distribution $\mathrm{p}(\boldsymbol{y}_i | \boldsymbol{x}_i, \boldsymbol{w})=\mathcal{N} \left( \mathbf{f}(\boldsymbol{x}_i,\boldsymbol{w}), \sigma^2 \right)$ for regression tasks and $\mathrm{p}(\boldsymbol{y}_i | \boldsymbol{x}_i, \boldsymbol{w})=\mathrm{Softmax}\left( \mathbf{f}(\boldsymbol{x}_i,\boldsymbol{w})\right)$ for classification problems.
By further modeling each neuron's weights as a mixture of two multivariate Gaussians with small variance $\sigma^2$ (the mean of one component set to $\boldsymbol{0}$ and the other mean $\boldsymbol{\theta}$) they show that minimizing the loss
\begin{equation}
L(\boldsymbol{\theta}) = -\sum_{i=1}^{N} \log(\mathrm{p}(\boldsymbol{y}_i | \mathbf{f}(\boldsymbol{x}_i, \hat{\boldsymbol{w}}_i)) + \frac{1 - p_{\mathrm{drop}}}{2N} \norm{\boldsymbol{\theta}}^2
\label{eq:logL}
\end{equation}
yields a reasonable\footnote{Minimizing $L(\boldsymbol{\theta})$ also minimizes the Kullback-Leiber divergence between $\mathrm{p}(\boldsymbol{w} | \boldsymbol{X}, \boldsymbol{Y})$ and the distribution $q(\boldsymbol{w})$ of all neuron's weights modeled as Gaussian mixtures, known as variational inference.} variational Bayesian approximation of $p(\boldsymbol{w} |\boldsymbol{X}, \boldsymbol{Y} )$.
Here, $\hat{\boldsymbol{w}}_i$ denote the masked model weights after applying dropout with probability $p_{\mathrm{drop}}$.
For regression problems the data term
\begin{equation}
\log \left( \mathrm{p} \left( \boldsymbol{y}_i | \mathbf{f} \left( \boldsymbol{x}_i, \hat{\boldsymbol{w}}_i \right) \right) \right) \propto \frac{\norm{\boldsymbol{y}_i - \mathbf{f}(\boldsymbol{x}_i, \hat{\boldsymbol{w}}_i))}^2}{\sigma^2} + \log(\sigma^2)
\label{eq:loss}
\end{equation}
can be expressed by squared data residuals and the noise parameter $\sigma$ (see \cite{Kendall2017}).
If $\sigma$ is fixed to a constant value, the problem reduces to a nonlinear least squares problem.
If $\sigma$ is assumed to be independent of the input and estimated, homoscedastic aleatoric uncertainty can be captured.
When $\sigma$ is modeled depending on the inputs, we capture heteroscedastic aleatoric uncertainty.
Kendall et al. \cite{Kendall2017} approximate the predictive mean and variance
\begin{equation}
\mathrm{E} \approx \frac{1}{T} \sum_{t=1}^T \hat{y}_t^*,~\mathrm{Var} \approx \frac{1}{T} \sum_{t=1}^T \left( \hat{y}_t^{*2} + \hat{\sigma}_t^2 \right) - \mathrm{E}^2
\label{eq:predictive_mean_and_variance}
\end{equation}
for each scalar $y^*$  with the regression output $\hat{y}^*_t$, $\sigma_t^2$ within $T$ runs with dropout applied during inference.
Note from Eq.~\ref{eq:predictive_mean_and_variance} that only one forward pass without dropout is necessary to estimate aleatoric uncertainties.

To estimate aleatoric classification uncertainties Kendall et al. \cite{Kendall2017} corrupt the scores $\mathbf{f} \left( \boldsymbol{x}^*, \boldsymbol{w} \right)$ with gaussian noise $\hat{\boldsymbol{y}} = \mathbf{f} \left( \boldsymbol{x}^*, \boldsymbol{w} \right) + \mathcal{N}\left(\boldsymbol{0}, \boldsymbol{\sigma}^2\right)$ with learned diagonal covariance $\boldsymbol{\sigma}^2$ and determine the softmax distribution $\hat{\boldsymbol{s}} = \mathrm{Softmax} \left( \hat{\boldsymbol{y}} \right)$. 
Finally, they modify the loss proposed in Eq.~\ref{eq:logL} to minimize the negative expected log likelihood
\begin{equation}
- \log \mathrm{E}(\hat{\boldsymbol{s}}) \approx \frac{1}{J} \sum_{j=1}^J \mathrm{H}_{\text{SCE}} \left(\boldsymbol{y},\hat{\boldsymbol{y}}_j\right)
\label{eq:loss_cls}
\end{equation}
with softmax-cross-entropy $\mathrm{H}_{\text{SCE}}$ using MC-integration.

\subsection{Uncertainties in Object Detection}
\label{sec:related_work_uncertainties_in_object_detection}

Feng et al. \cite{Feng2018} present a vehicle detector that estimates epistemic and aleatoric box regression as well as epistemic classification uncertainties.
The authors use a Resnet8 feature extractor on a modified Faster-RCNN meta architecture.
A conventional region proposal network (RPN) provides regions of interest (ROIs) that serve as input for three subsequent fully-connected layers with dropout producing classification and bounding box regression, respectively.
3D bounding boxes are encoded by eight corner points, resulting in 24 parameters for each box.
The authors evaluate their framework on the KITTI raw dataset which provides subsequent lidar measurements.
Feng et al. show that epistemic uncertainty correlates to detection accuracy.
In contrast, they find that aleatoric uncertainty correlates to the number of measurements for each object by comparing fully visible to distant or partially occluded objects.
More recently, in \cite{Feng2018a} the same authors improve average precision and runtime by only estimating aleatoric uncertainty.

Based on the work of Feng et al. \cite{Feng2018, Feng2018a} we add the detection of cyclists and pedestrians.
Furthermore, we validate uncertainty estimation during the RPN and the output stage for different models and analyze their differences.
We evaluate our work on the KITTI Bird's Eye View Evaluation benchmark allowing comparison to previously published work \cite{Wirges2018}.
Finally, we reduce the number of box parameters and represent pose and shape uncertainty by a common shape based on collision probabilities.
This representation may be helpful for subsequent trajectory planning, e.g. as presented in \cite{Banzhaf2018}.

\subsection{KITTI Bird's Eye View Benchmark}
\label{sec:related_work_kitti_bev_benchmark}

The KITTI Bird's Eye View (BEV) Evaluation 2017 \cite{Geiger2012} provides sensor measurements together with semantically and spatially annotated 3D bounding boxes for relevant traffic participants.
It consists of 7481 training and 7518 testing samples comprising a total number of 80,256 labeled objects.
Samples consist of camera images and range sensor data, represented as point sets.
There are separate benchmarks available for cars, cyclists and pedestrians.
The authors use average precision as evaluation metric which is determined similar to the PASCAL protocol by averaging over 10\% recall steps.

Currently, most of the best performing benchmark submissions rely only on range sensor measurements and design their networks to detect one object class \cite{Yang2018, Du2018, Shi2018}.
Yang et al. \cite{Yang2018} use either offline maps or inferred maps as prior information to increase object detection performance.
Du et al. \cite{Du2018} segment range sensor measurements by object detections from camera images and perform shape estimation on the segmented measurements.
Shi et al. \cite{Shi2018} generate proposals directly from the 3D point set domain and subsequently estimate refined boxes from canonical coordinates.

\section{Model}
\label{sec:model}

\subsection{Inputs}
\label{sec:model_inputs}
\begin{figure}[ht]
\centering
\begin{subfigure}{0.24\linewidth}
\includegraphics[width=\linewidth]{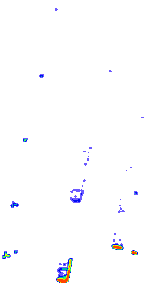}
\caption{Detections}
\label{fig:detections}
\end{subfigure}
\begin{subfigure}{0.24\linewidth}
\includegraphics[width=\linewidth]{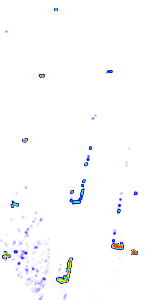}
\caption{Height diff.}
\label{fig:height_difference}
\end{subfigure}
\begin{subfigure}{0.24\linewidth}
\includegraphics[width=\linewidth]{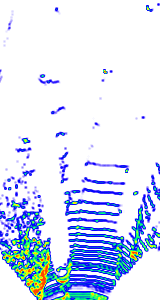}
\caption{Intensity}
\label{fig:intensity}
\end{subfigure}
\begin{subfigure}{0.24\linewidth}
\includegraphics[width=\linewidth]{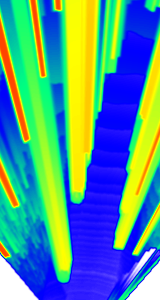}
\caption{Occ. height}
\label{fig:occlusion_height}
\end{subfigure}

\caption{
Grid map layers used.
White/blue indicates low, red high values.
The number of detections per cell depend on distance and structure verticality.
Height information above the ground surface is expressed by the min. / max. z coordinate layers (here: height difference).
The intensity layer carries information on the surface material.
In contrast to the detection height, the occlusion height encodes the amount of observable space above ground.
}
\label{fig:layers}
\end{figure}

We use a multi-layer top-view grid map as input data for our object detector with the layers depicted in Fig.~\ref{fig:layers} which we found particularly suitable for fusion of multiple range sensors.
However, here our grid maps are constructed from only one range sensor (available in KITTI).
Compared to previous work \cite{Wirges2018} we use the maximum height of occlusions above ground casted by objects as an additional layer (see Fig.~\ref{fig:occlusion_height}).
We found this layer helpful for fusion tasks and its features seem to be discriminant with respect to the different traffic participants.

In contrast to Feng et al. \cite{Feng2018} we do not encode the height of detections in several layers but instead represent occupied space by its range encoded into two layers (illustrated by Fig.~\ref{fig:height_difference}).

\subsection{Outputs}
\label{sec:model_outputs}
Each object is assigned a semantic class, pose, shape (represented as bounding box) and uncertainties.
The model estimates mean and variance of the proposal-relative regression parameters
\begin{equation}
\boldsymbol{y} = \left[ x, y, \log \left( b_{\mathrm{l}} \right), \log \left( b_{\mathrm{w}} \right), \sin \left( 2 \phi \right), \cos \left( 2 \phi \right) \right]
\label{eq:parameters}
\end{equation}
which we assume to be normally distributed and uncorrelated.
Here, $x$ and $y$ denote the object position, $b_{\mathrm{l}}$ and $b_{\mathrm{w}}$ the box length and width and $\phi$ the object rotation.
Note that box length $b_{\mathrm{l}}$ and width $b_{\mathrm{w}}$ follow a log-normal distribution and that because of $\sin \left( 2 \phi \right)$ and $\cos \left( 2 \phi \right)$ an analytical distribution of $\phi$ is intractable.

Although expressing uncertainty is fully explained in the object parameter space subsequent trajectory planning stages may require a more condensed shape representation.
Therefore, we develop an approach to map objects from a six-dimensional representation to a volumetirc representation in the grid map frame.
Based on lower and upper percentiles for $\sin(2\phi)$ and $\cos(2\phi)$ we sample rotations equidistantly between $\phi_{\mathrm{min}} = \frac{1}{2} \tan^{-1} \left( \frac{\sin(2\phi)_{\mathrm{min}}}{\cos(2\phi)_{\mathrm{max}}} \right)$ and $\phi_{\mathrm{max}} = \frac{1}{2} \tan^{-1} \left( \frac{\sin(2\phi)_{\mathrm{max}}}{\cos(2\phi)_{\mathrm{min}}} \right)$.
To determine the box face distributions we transform position variances $\sigma^2_{\mathrm{x}}$ and $\sigma^2_{\mathrm{y}}$ into the box coordinate frame.
Since the convolution between normal and log-normal distribution cannot be evaluated analytically, we use a Monte-Carlo approach to estimate the face distributions.
From the resulting histograms $\hat{b}_{\mathrm{l}}$ and $\hat{b}_{\mathrm{w}}$ we determine an upper percentile, determine the corner points and transform them back into the sensor frame.
Finally, we compute the convex hull as a compact representation and conservative over-approximation of the true shape distribution.
Exemplary convex hulls are depicted in Fig.~\ref{fig:top_view} and Tab.~\ref{tab:qual_objects}.

\subsection{Network Structure and Training Process}
\label{sec:model_network_structure_and_training_process}
Starting from a previously published baseline model (model 4 in \cite{Wirges2018}) we focus only on the modifications made for the different configurations depicted in Table~\ref{tab:benchmark_results}.
To estimate only epistemic uncertainty we add an additional dropout layer with $p_{\mathrm{drop}} = 0.2$ to the baseline's head right before its three fully-connected layers, referred to as \textit{Dropout Head}.
To apply dropout to the convolutional layers we use the method proposed in \cite{Gal2015b} which drops out feature map values before activation.
We also build a model with dropout $p_{\text{drop}} = 0.2$ in every convolutional and fully-connected layer, referred to as \textit{Dropout Fully} and a model which only applies dropout to all layers of the second stage, \textit{Conv. Dropout}.
In order to estimate epistemic uncertainty according to Eq.~\ref{eq:predictive_mean_and_variance} we compare three different approaches.
In the first one we use \textit{Conv. Dropout} and pass the same multi-layer grid map $T$-times through the network. 
To reduce computational time we copy the final feature map given by the feature-extractor $T$-times in \textit{Dropout Head} and subsequently process it as batch.
In the third approach we also use \textit{Dropout Head} and batches but instead of calculating mean and variance after postprocessing we compute them directly after the final layers and thus before non-maxima-suppression. 
This approach only needs matrix operations and can further decrease inference time. 
Because we calculate uncertainties for each region proposal here we refer to it as \textit{Anch. Dropout}.
To predict aleatoric regression uncertainties we add additional layers to the head as proposed by \cite{Kendall2017} and minimize the loss presented in Eq.~\ref{eq:loss}.
This results in an L2-Loss and we refer to it as \textit{Aleatoric L2}.
To analyze the influence of the loss function we further use a variant which uses an L1-Loss (as presented in \cite{Feng2018}) instead of an L2-Loss, named  \textit{Aleatoric L1}.
In order to predict epistemic and aleatoric uncertainties, we combine \textit{Dropout Head} and \textit{Aleatoric L1} in configuration \textit{L1 \& Dropout}.
In order to investigate the influence of estimating aleatoric classification uncertainties on network's performance we also construct a model which makes predictions according to Eq.~\ref{eq:loss_cls} in the head (named \textit{Aleatoric Cls.}) and in head and region proposal network (named \textit{Aleatoric Cls. + RPN Cls.}).

In exactly the same way as \cite{Wirges2018}, all models are trained on a 69\% / 31\% KITTI BEV training / validation set with the same training parameters.
However, as our modifications have an impact on the convergence we train our models for different epochs.
We first train the baseline model for 352 epochs without uncertainty estimation. 
Then we train each network for another 230 epochs, gradually reducing the learning rate from $\mathrm{5 \cdot 10^{-4}}$ to $\mathrm{1 \cdot 10^{-6}}$.
To mitigate overfitting we validate every 48 epochs and perform early-stopping.

\section{Evaluation}
\label{sec:evaluation}

\subsection{Metrics}
\label{sec:evaluation_metrics}
As described in Sec.~\ref{sec:related_work_modeling_uncertainty_in_neural_networks} the variance for each network output can be directly interpreted as parameter uncertainty.
Here we use two different coordinate systems to represent the parameters, coordinates normalized by the grid map size and KITTI coordinates in m. 
Because we estimate $\sigma_{\log \left(b_{\mathrm{l}} \right)}$ and $\sigma_{\log \left(b_{\mathrm{w}} \right)}$, we apply exponential mapping $\exp \left( \sigma_{\log \left(b_{\mathrm{l}} \right)} \right)$ and $\exp \left( \sigma_{\log \left(b_{\mathrm{w}} \right)} \right)$ to better represent those uncertainties.
If $b_{\mathrm{l}}$ and $b_{\mathrm{w}}$ are assumed to be log-normal distributed this is equal to the multiplicative standard deviation which implies $p \left( b_{\mathrm{l},0.5} \cdot \exp \left( \sigma_{\log \left( b_{\mathrm{l}} \right)} \right)\right) = 0.68$.
To describe the overall regression uncertainties of an object in one single value the total variance (TV) of the parameters $\boldsymbol{y}$ in Eq.~\ref{eq:parameters} is used. 
Since the parameter units are not consistent the TV cannot be interpreted physically.
Nevertheless it is helpful to compare different objects.
To describe epistemic classification uncertainties we use MC-integrated Shannon-entropy \cite{Gal2016} $\mathrm{H} \left(y^*|\boldsymbol{x}^*,\boldsymbol{X},\boldsymbol{Y}\right) = -\sum_{c} \overline{s}_{c} \log \overline{s}_{c}$ for each object with $\overline{s}_{c}= \frac{1}{T} \sum_{t=1}^T \text{Softmax}\left(\boldsymbol{\mathrm{f}}_{\hat {\boldsymbol{w}}}(\boldsymbol{x}^*)\right)_{c}$ indicating the mean softmax score for $T$ runs during inference for class $c$.

\subsection{Qualitative Results}
\label{sec:evaluation_qualitative_results}

\begin{table*}[ht]
\setlength{\tabcolsep}{3pt}
\centering
\begin{tabular}{ccccccc}
\toprule
Dropout Head & Conv. Dropout & Anch. Dropout & Aleatoric L1 & Aleatoric L2 & Aleatoric L1 \& Dropout Head \\
\raisebox{-0.9\totalheight}{\includegraphics[height=1.5cm]{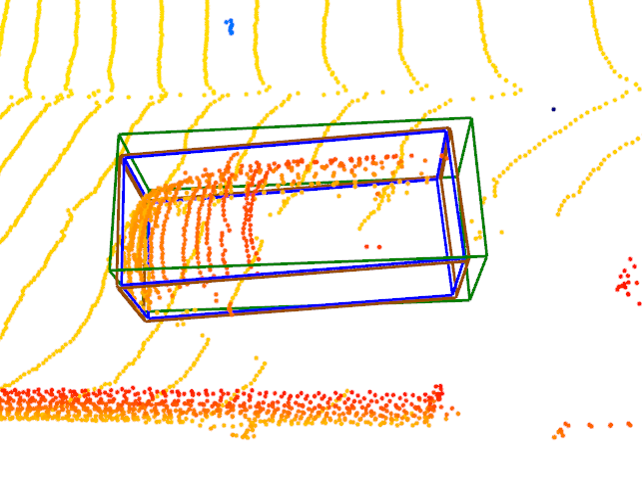}} & 
\raisebox{-0.9\totalheight}{\includegraphics[height=1.5cm]{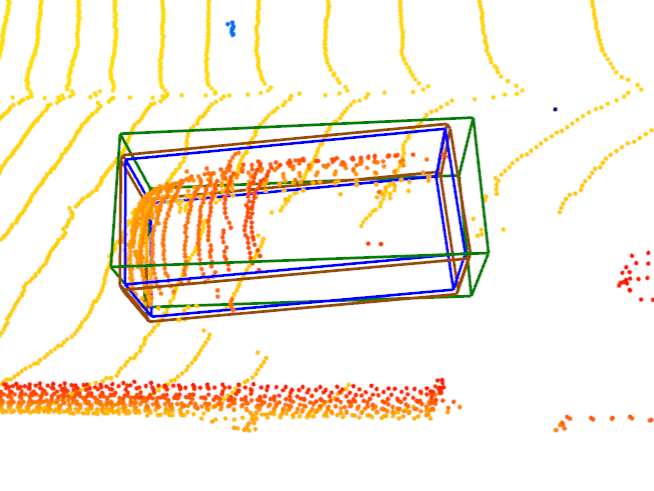}}  & 
\raisebox{-0.9\totalheight}{\includegraphics[height=1.5cm]{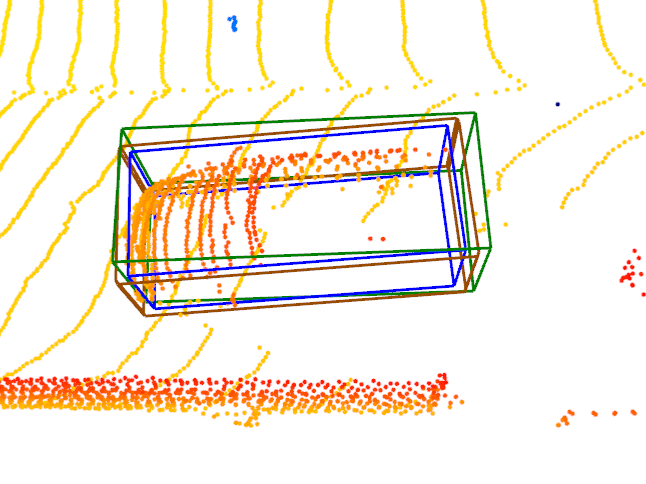}}  & 
\raisebox{-0.9\totalheight}{\includegraphics[height=1.5cm]{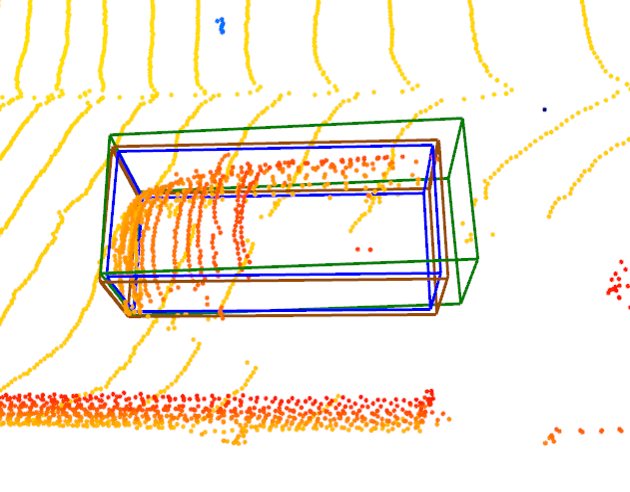}}  & 
\raisebox{-0.9\totalheight}{\includegraphics[height=1.5cm]{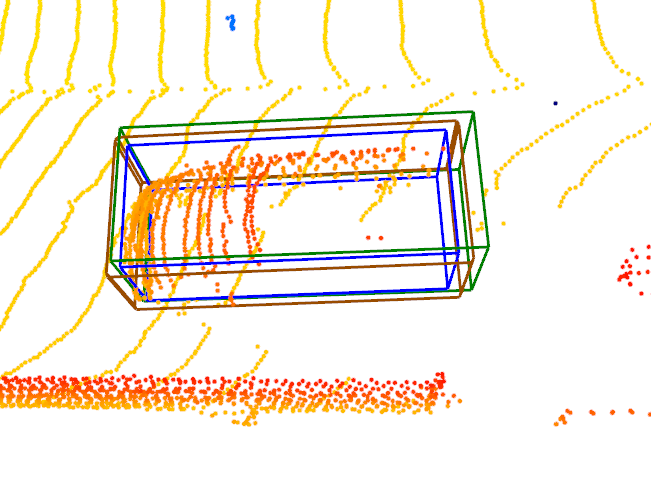}}  &
\raisebox{-0.9\totalheight}{\includegraphics[height=1.5cm]{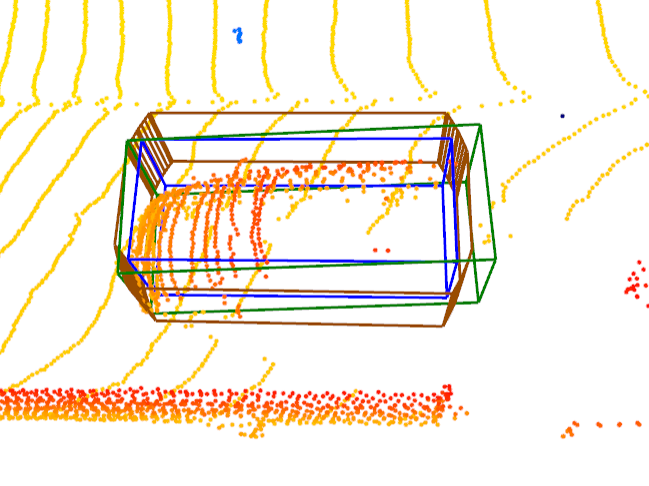}}  \\
TV: $\num{0e-4}$ & TV: $\num{2e-4}$ & TV: $\num{35e-4}$ & TV: $\num{1e-4}$ & TV: $\num{40e-4}$ & TV: $\num{27e-4}$ \\
\midrule
Anch. Dropout & Anch. Dropout & Aleatoric L2 & Aleatoric L2 & L1 \& Dropout Head & L1 \& Dropout Head \\
\raisebox{-0.9\totalheight}{\includegraphics[height=1.5cm]{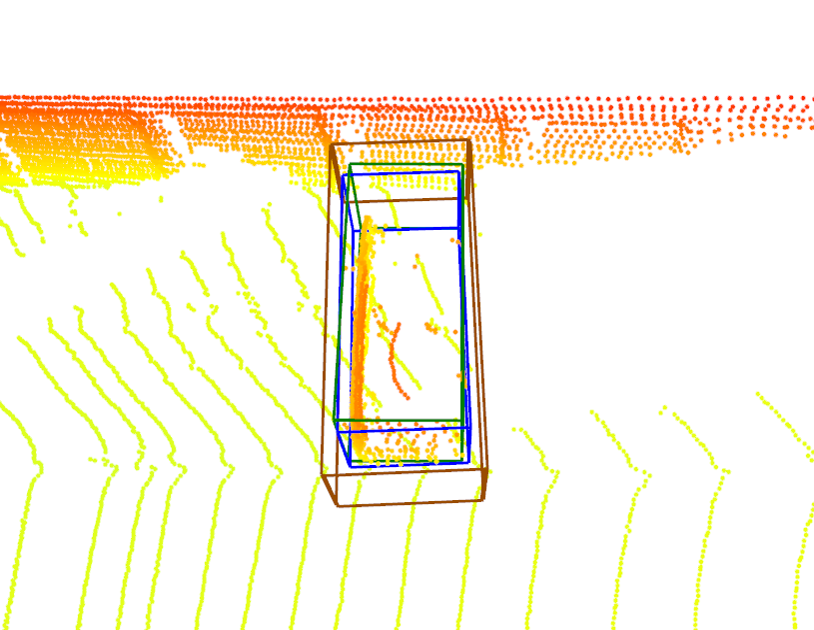}} & 
\raisebox{-0.9\totalheight}{\includegraphics[height=1.5cm]{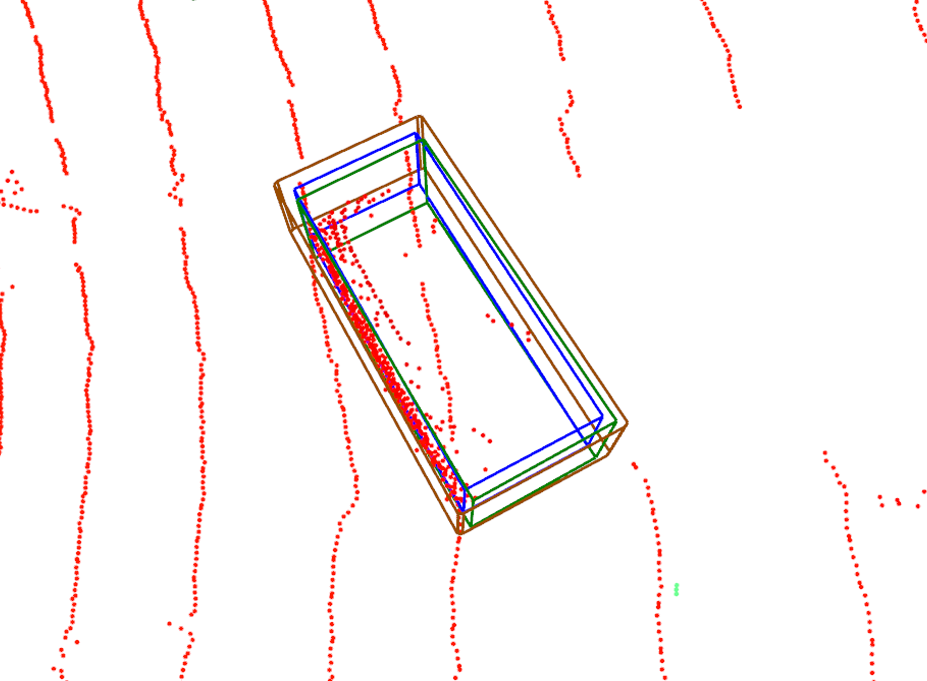}}  & 
\raisebox{-0.9\totalheight}{\includegraphics[height=1.5cm]{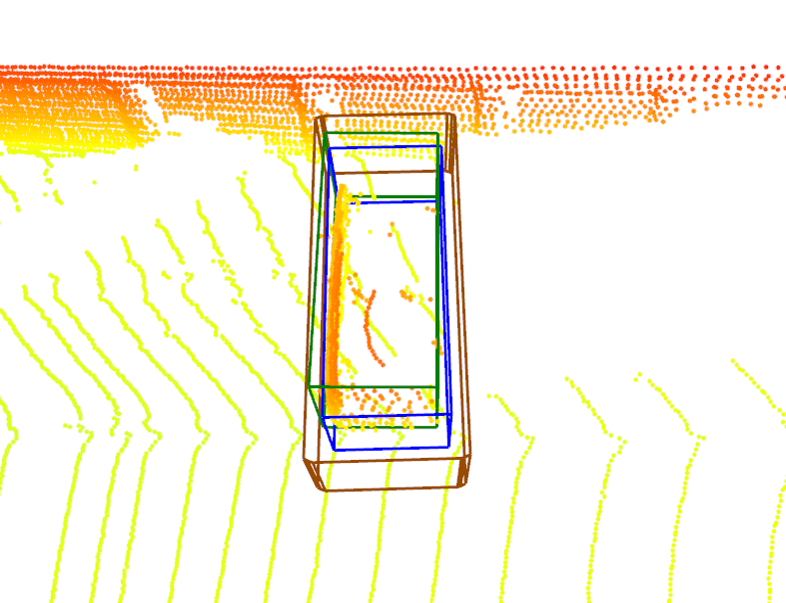}}  & 
\raisebox{-0.9\totalheight}{\includegraphics[height=1.5cm]{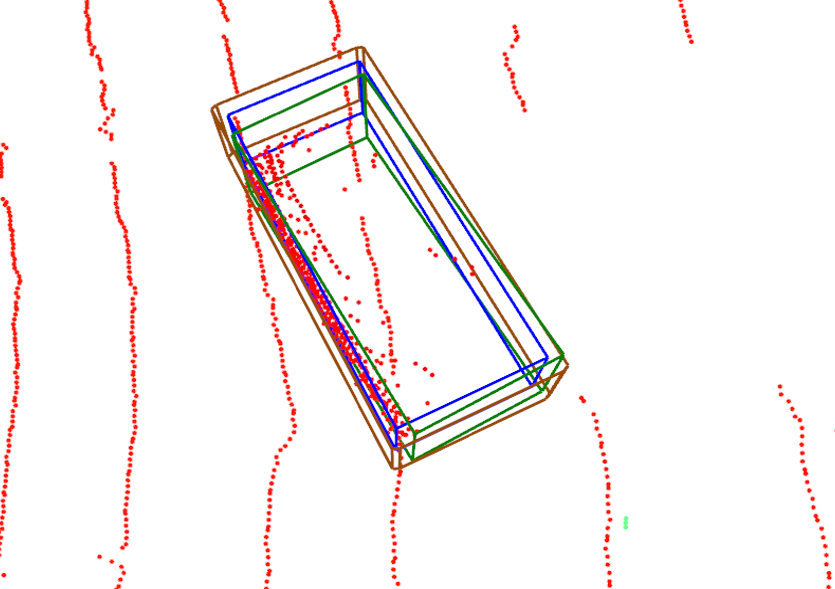}}  & 
\raisebox{-0.9\totalheight}{\includegraphics[height=1.5cm]{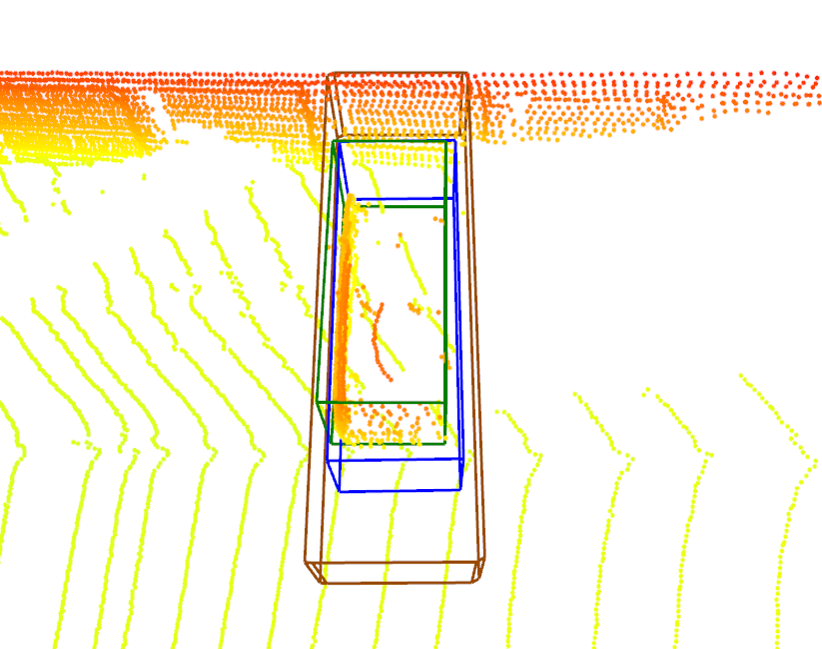}}  &
\raisebox{-0.9\totalheight}{\includegraphics[height=1.5cm]{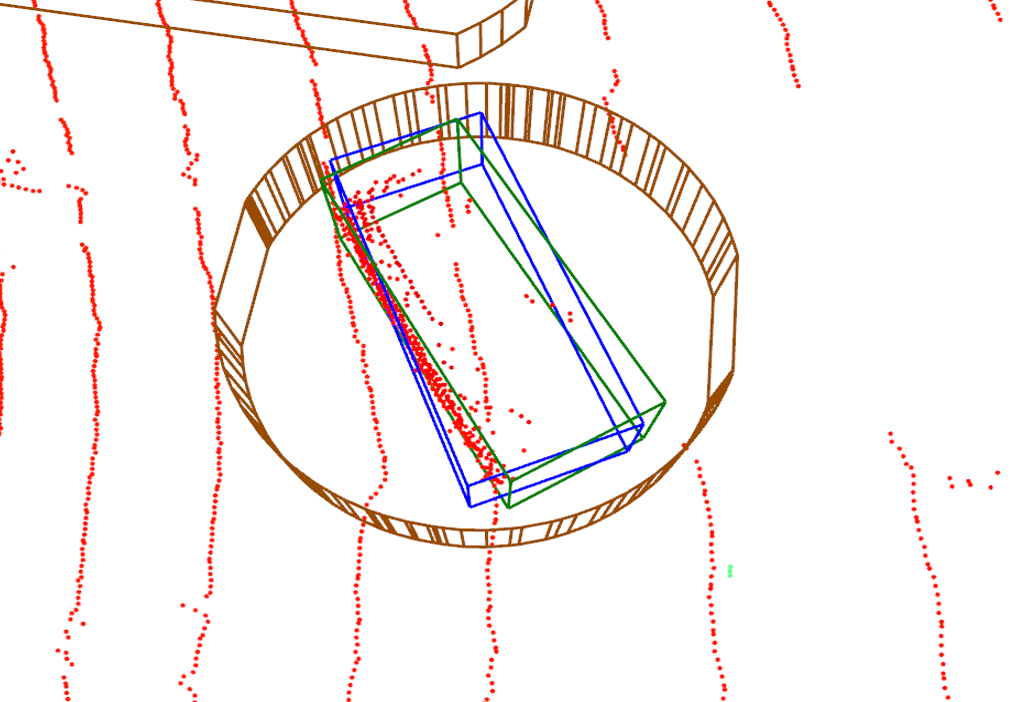}}  \\
TV: $\num{336e-4}$ & TV: $\num{129e-4}$ & TV: $\num{290e-4}$ & TV: $\num{46e-4}$ & TV: $\num{738e-4}$  & TV: $\num{851e-4}$ \\
\midrule
Anch. Dropout & Aleatoric L2 & L1 \& Dropout Head & Anch. Dropout &  Aleatoric L2 & L1 \& Dropout Head \\
\raisebox{-0.9\totalheight}{\includegraphics[height=1.5cm]{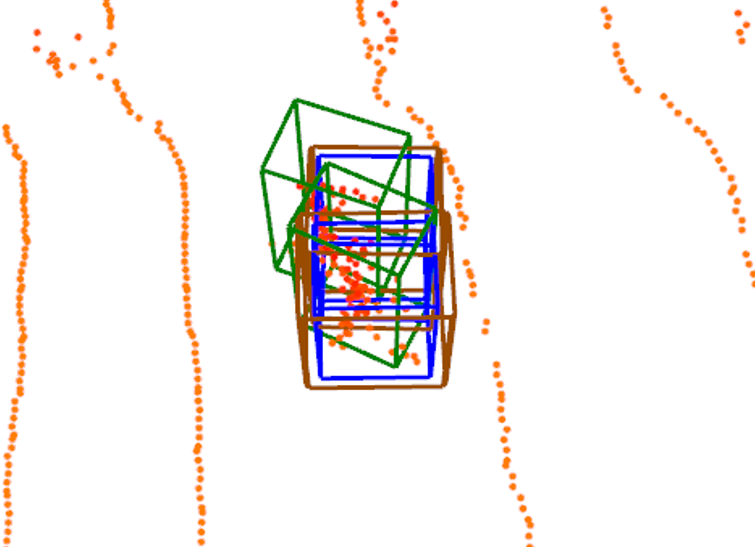}} & 
\raisebox{-0.9\totalheight}{\includegraphics[height=1.5cm]{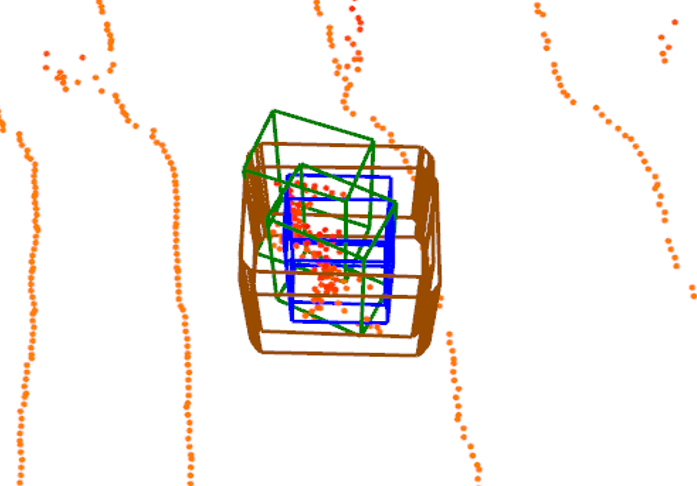}}  & 
\raisebox{-0.9\totalheight}{\includegraphics[height=1.5cm]{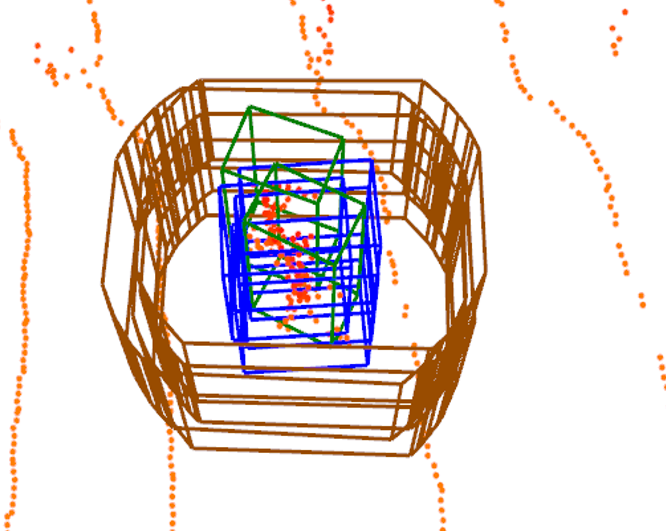}}  & 
\raisebox{-0.9\totalheight}{\includegraphics[height=1.5cm]{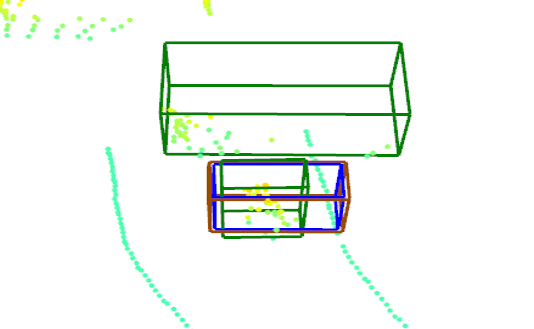}}  &
\raisebox{-0.9\totalheight}{\includegraphics[height=1.5cm]{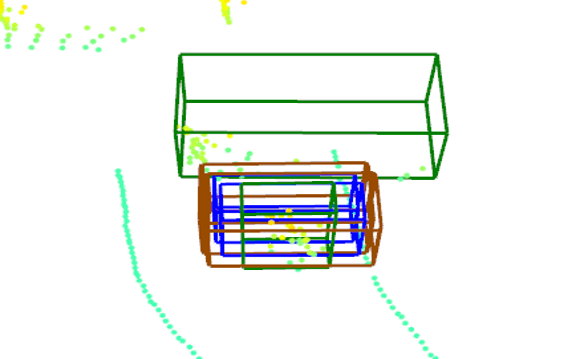}} & 
\raisebox{-0.9\totalheight}{\includegraphics[height=1.5cm]{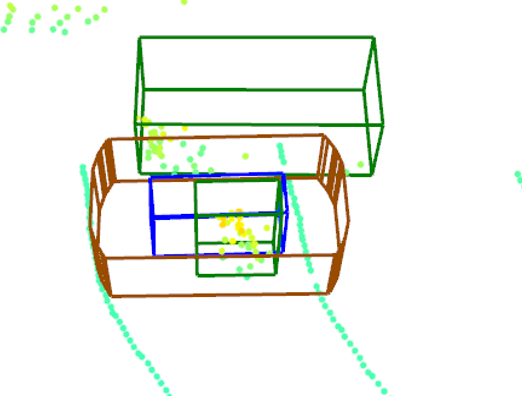}}  \\
TV: $\num{265e-4}$ & TV: $\num{2134e-4}$ &  TV: $\num{3864e-4}$ &  TV: $\num{50e-4}$ & TV: $\num{411e-4}$ & TV: $\num{3337e-4}$  \\
\bottomrule
\end{tabular}
\caption{
Qualitative results of uncertainty estimation for specific objects.
The Median-Bounding-Box is given in blue, the convex hull for the 95th percentile in red and the ground truth in green.
TV is given in normalized grid map coordinates.
The first row shows results for one specific car for all presented architectures.
The second row contains results for an object that is oriented orthogonal to the direction of travel and an object that is oriented under $45^\circ$.
The third row shows specific results for pedestrians and cyclists.}
\label{tab:qual_objects}
\end{table*}

Fig.~\ref{tab:qual_objects} shows that epistemic uncertainties seem to be very small compared to aleatoric uncertainties which are for cars practically identical in most cases. 
Because for inaccurately estimated objects in many cases convex hulls do not envelop the ground truth bounding box, we think that the chosen approach underestimates epistemic uncertainties. 
This effect ist known from \cite{Li2017}.
Aleatoric uncertainty estimation seems to be more realistic. 
Combined uncertainty estimation causes highest uncertainties, which can be also seen in its poor accuracy (see Tab.~\ref{tab:benchmark_results}) and thus higher model uncertainties.

\subsection{Quantitative Results}
\label{sec:evaluation_quantitative_results}

\subsubsection*{KITTI BEV benchmark}
\readcsv{results/performance/best_results_val.csv}{\benchmarkvalresults}
\readcsv{results/performance/best_results_test.csv}{\benchmarktestresults}
\pgfplotstablevertcat{\benchmarkresults}{\benchmarktestresults}
\pgfplotstablevertcat{\benchmarkresults}{\benchmarkvalresults}
\pgfplotstableset{
columns/step/.style={column name=Epoch, precision=0, preproc/expr={8/5150*##1}, column type={c|}},
columns/car_easy/.style={column name=Car Easy, multiply by={100}, column type={c}},
columns/car_moderate/.style={column name=Car, multiply by={100}, column type={c}},
columns/car_hard/.style={column name=Car Hard, multiply by={100}, column type={c}},
columns/pedestrian_easy/.style={column name=Pedestrian Easy, multiply by={100}, column type={c}},    
columns/pedestrian_moderate/.style={column name=Pedestrian, multiply by={100}, column type={c}},
columns/pedestrian_hard/.style={column name=Pedestrian Hard, multiply by={100}, column type={c}},
columns/cyclist_easy/.style={column name=Cyclist Easy, multiply by={100}, column type={c}},
columns/cyclist_moderate/.style={column name=Cyclist, multiply by={100}, column type={c}},
columns/cyclist_hard/.style={column name=Cyclist Hard, multiply by={100}, column type={c}},
columns/time/.style={column name=Time / ms, multiply by={1000}, column type={c}},
}

\pgfplotstableset{
numeric type,
fixed,
zerofill,
precision=1,
every first column/.style={column type/.add={|}{}},
every last column/.style={column type/.add={}{|}},
every last row/.style={after row=\bottomrule},	
}

\pgfplotstableset{
create on use/experimentsname/.style={
create col/set list={Baseline,Aleatoric Cls.,$\quad+\text{RPN,Cls.}$,Aleatoric L1,Baseline,Dropout Head,$\quad$Ep,$\quad$Anch. Dropout,$\text{Conv. Dropout}$,$\quad$Ep,Aleatoric Cls.,$\quad+\text{RPN,Cls.}$,$\quad+\text{RPN,Cls.}$,$\quad+\text{RPN,Cls.}$,Aleatoric L1,Aleatoric L1,Aleatoric L2, Combined,$\quad$Ep}}, columns/experimentsname/.style={string type, column name=Experiment, column type={l}}}

\begin{table}[!ht]
\centering
\setlength{\tabcolsep}{3pt}

\pgfplotstabletypeset[
every head row/.style={before row=\toprule, after row=\midrule}, every row no 3/.style={after row=\midrule},
columns={experimentsname,step,car_moderate,pedestrian_moderate,cyclist_moderate,time},
columns/experiment/.style={string type, column name=Experiment, column type={c|}},
every row 0 column 2/.style={postproc cell content/.style={@cell content/.add={$\bf}{$}}},
every row 4 column 2/.style={postproc cell content/.style={@cell content/.add={$\bf}{$}}},
every row 2 column 3/.style={postproc cell content/.style={@cell content/.add={$\bf}{$}}},
every row 11 column 3/.style={postproc cell content/.style={@cell content/.add={$\bf}{$}}},
every row 0 column 4/.style={postproc cell content/.style={@cell content/.add={$\bf}{$}}},
every row 14 column 4/.style={postproc cell content/.style={@cell content/.add={$\bf}{$}}},
]{\benchmarkresults}
	
\caption{
Results on KITTI test and validation set.
We show the average precision for car, pedestrian and cyclist in moderate difficulty and inference times.
Ep indicates that epistemic uncertainty is estimated during inference.
Accuracy strongly depends on trained epochs.
}
\label{tab:benchmark_results}
\end{table}

Tab.~\ref{tab:benchmark_results} summarizes the results of our proposed models on KITTI test and validation set.
Applying dropout to the final layers seems to improve accuracy for pedestrians whereas accuracy for car suffers. 
\textit{Dropout Fully} does not seem to converge to a reasonable solution, which we think is because of great information loss in the first layers of the feature extractor.
The accuracy for the three classes behaves different depending on the number of trained epochs, indicating that overfitting for car, pedestrian and cyclists starts at different epochs.
As expected, using batches reduces inference time for estimating epistemic uncertainty (see \textit{Conv. Dropout} and \textit{Dropout Head}). 
Calculating mean and variance per region proposal further reduces inference times and seems to be suitable for real time applications as using $15$ forwards passes comes only with $30.2$ms additional computation time overhead.

\subsubsection*{Number of forward passes}
\readcsv{results/forward_passes/dropout_p02/total_var_boxes_over_t.csv}{\dropoutfiltboxesovert}
\readcsv{results/forward_passes/dropout_p02/total_var_scores_over_t.csv}{\dropoutfiltscoresovert}
\readcsv{results/forward_passes/conv_dropout/total_var_boxes_over_t.csv}{\convdropoutfiltboxesovert}
\readcsv{results/forward_passes/conv_dropout/total_var_scores_over_t.csv}{\convdropoutfiltscoresovert}

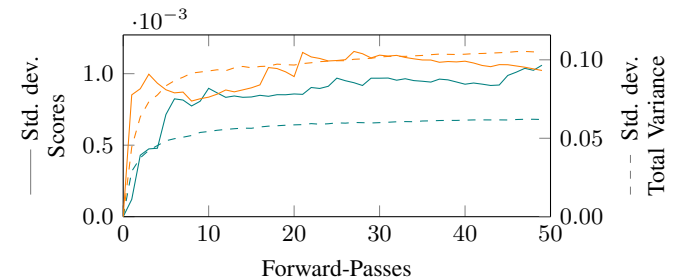
\begin{figure}[ht]
\centering

\begin{tikzpicture}
\tikzstyle{every node}=[font=\small]
\begin{axis}[
width=0.82\columnwidth,
height=4cm,
xmin=0,
xmax=50,
ymin=0,
xlabel={Forward-Passes},
xticklabel style={/pgf/number format/precision=0},
axis y line*=left,
ylabel style={align=center},
ylabel={\ref{pgfplots:plot1} Std. dev. \\ Scores},
yticklabel style={/pgf/number format/fixed, /pgf/number format/precision=1},
]
\addplot+[mark=None] table [y index=1]{\dropoutfiltscoresovert};
\addplot+[mark=None] table [y index=1]{\convdropoutfiltscoresovert};
\addlegendimage{no markers, gray}
\label{pgfplots:plot1}
\end{axis}

\begin{axis}[
width=0.82\columnwidth,
height=4cm,
xmin=0,
xmax=50,
ymin=0,
axis y line*=right,
axis x line=none,
ylabel style={align=center},
ylabel={\ref{pgfplots:plot2} Std. dev. \\ Total Variance},
yticklabel style={/pgf/number format/fixed, /pgf/number format/precision=2},
]
\addplot+[mark=None, dashed] table [y expr={\thisrowno{1}+\thisrowno{2}+\thisrowno{3}+\thisrowno{4}+\thisrowno{5}+\thisrowno{6}}]{\dropoutfiltboxesovert};
\addplot+[mark=None, dashed] table [y expr={\thisrowno{1}+\thisrowno{2}+\thisrowno{3}+\thisrowno{4}+\thisrowno{5}+\thisrowno{6}}]{\convdropoutfiltboxesovert};
\addlegendimage{no markers, dashed, gray}
\label{pgfplots:plot2}
\end{axis}
\end{tikzpicture}

\caption{
Standard deviations over stochastic forward passes during inference.
Empirical standard deviations of softmax classification scores and standard deviations of total variance for all box parameters (in normalized grid-map coordinates) depending on number of forward passes $T$ for calculating predictive variance according to Eq.~\ref{eq:predictive_mean_and_variance}.
Dark: \textit{Dropout Head}, bright: \textit{Conv. Dropout}.
}
\label{fig:stddev_over_t}
\end{figure}

To choose a reasonable number of forward runs $T$ to determine epistemic uncertainties we calculate the standard deviation of scores and the total variance of 100 boxes after every forward run and sum them up.
As can been seen from Fig.~\ref{fig:stddev_over_t} this converges after approx. ten runs.
In the following we use $T=15$.
As can be seen from Fig.~\ref{fig:stddev_over_t} as well as from Fig.~\ref{fig:unc_over_iou} and \ref{fig:epistemic_unc_over_dist} \textit{Conv. Dropout} causes higher uncertainties than \textit{Dropout Head}.
This confirms Gal et al. \cite{Gal2015b} who suggest that dropout should be applied to all convolutional layers.

\subsubsection*{Validation}
\readcsv{results/epistemic_unc/dropout_p02_on_baseline_123600/entropy_over_iou.csv}{\dropoutentroiou}
\readcsv{results/epistemic_unc/dropout_p02_on_baseline_123600/tv_over_iou.csv}{\dropouttvoiou}
\readcsv{results/epistemic_unc/dropout_p02_on_baseline_123600/stddev_over_iou.csv}{\dropoutstddevoiou}
\readcsv{results/epistemic_unc/dropout_p02_conv_second_stage_on_baseline_123600/entropy_over_iou.csv}{\convdropoutentroiou}
\readcsv{results/epistemic_unc/dropout_p02_conv_second_stage_on_baseline_123600/tv_over_iou.csv}{\convdropouttvoiou}
\readcsv{results/epistemic_unc/dropout_p02_conv_second_stage_on_baseline_123600/stddev_over_iou.csv}{\convdropoutstddevoiou}
\readcsv{results/epistemic_unc/dropout_p02_on_baseline_123600_anchorwise/entropy_over_iou.csv}{\anchdropoutentroiou}
\readcsv{results/epistemic_unc/dropout_p02_on_baseline_123600_anchorwise/tv_over_iou.csv}{\anchdropouttvoiou}
\readcsv{results/epistemic_unc/dropout_p02_on_baseline_123600_anchorwise/stddev_over_iou.csv}{\anchdropoutstddevoiou}
\readcsv{results/epistemic_unc/dropout_p02_on_baseline_123600/iou_hist.csv}{\dropoutiouhist}
\readcsv{results/aleatoric_unc/aleatoric_loc_381100/entropy_over_iou.csv}{\locentroiou}
\readcsv{results/aleatoric_unc/aleatoric_loc_381100/tv_over_iou.csv}{\loctvoiou}
\readcsv{results/aleatoric_unc/aleatoric_loc_381100/stddev_over_iou.csv}{\locstddevoiou}
\readcsv{results/aleatoric_unc/aleatoric_loc_l2_unc_init_185400/entropy_over_iou.csv}{\locltwoentroiou}
\readcsv{results/aleatoric_unc/aleatoric_loc_l2_unc_init_185400/tv_over_iou.csv}{\locltwotvoiou}
\readcsv{results/aleatoric_unc/aleatoric_loc_l2_unc_init_185400/stddev_over_iou.csv}{\locltwostddevoiou}

\begin{figure}[ht]
\centering

\begin{tikzpicture}
\tikzstyle{every node}=[font=\small]
\begin{groupplot}[group style={group size=1 by 4, vertical sep=1mm}]

\nextgroupplot[
xmax=1,
xticklabels={,,},
ytick distance=,
ylabel style={align=center},
ylabel={Shannon-Entropy \\ (epistemic)},
legend pos = north east,
]
\addplot table [y index=1]{\dropoutentroiou};
\addlegendentry{Dropout};
\addplot table [y index=1]{\convdropoutentroiou};
\addlegendentry{Conv. Dropout};
\addplot table [y index=1]{\anchdropoutentroiou};
\addlegendentry{Anch. Dropout};

\nextgroupplot[
xmax=1,
ytick distance=,
ymax=0.5,
legend pos = north east,
xticklabels={,,},
ylabel style={align=center},
ylabel={Total Variance \\ (epistemic)},
]
\addplot table [y index=1]{\dropouttvoiou};
\addlegendentry{Dropout};
\addplot table [y index=1]{\convdropouttvoiou};
\addlegendentry{Conv. Dropout};
\addplot table [y index=1]{\anchdropouttvoiou};
\addlegendentry{Anch. Dropout};

\nextgroupplot[
xmax=1,
ymax=,
ytick distance=,
ylabel style={align=center},
ylabel={Total Variance \\ (aleatoric)},
xticklabels={,,},
]
\pgfplotsset{cycle list shift=3}
\addplot table [y index=1]{\loctvoiou};
\addlegendentry{Aleatoric L1};
\addplot table [y index=1]{\locltwotvoiou};
\addlegendentry{Aleatoric L2};

\nextgroupplot[
ymin=1,
xmax=1,
xlabel={IoU},
ymode=log,
ytick={1, 10, 100, 1000},
ylabel={Abs. Frequency},
area style,
]
\addplot+[ybar interval,mark=no,teal!80!black,fill=teal!80!black,fill=teal!80!black,fill=teal!80!black, opacity=0.8] table [y index=1]{\dropoutiouhist};
\end{groupplot}
\end{tikzpicture}

\caption{
Epistemic class/regression and aleatoric regression uncertainties over IoU for different configurations in KITTI coordinates.
Low uncertainties in range $[0,0.1]$ can be explained by false-positives as can be seen in the histogram.
}
\label{fig:unc_over_iou}
\end{figure}
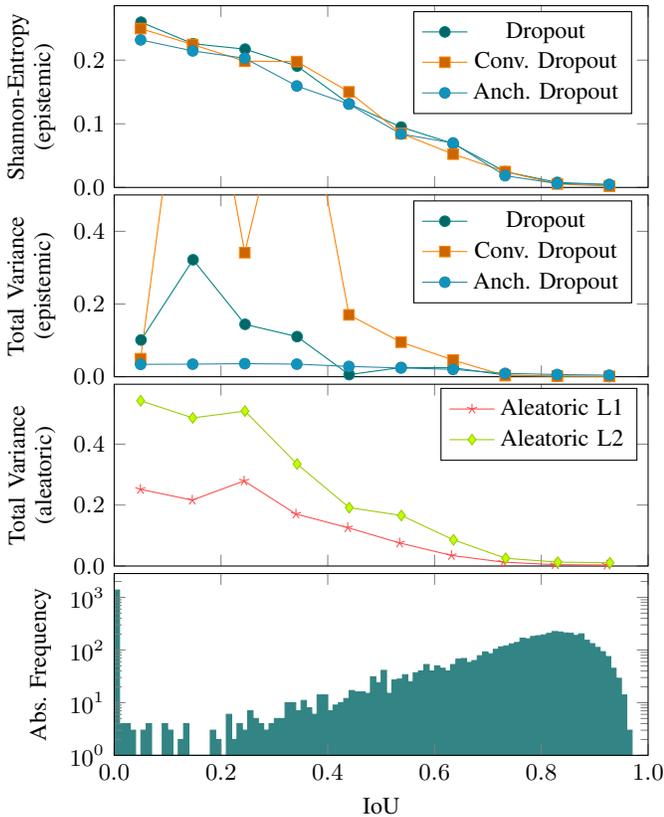

To validate epistemic and aleatoric uncertainty estimation we compare uncertainties over the intersection over union (IoU) in Fig.~\ref{fig:unc_over_iou}.
For this purpose we calculate the mean of all objects over 10\% steps.
In the case of IoU > 0.1 we observe uncertainties decrease with increasing IoU. 
If IoU < 0.1 epistemic regressions uncertainties are strongly influenced by false-positives which not seem to have higher model uncertainties than inaccurately estimated objects.
For aleatoric uncertainties this effect is not prevalent, indicating that false-positives have very high sensor uncertainties.
Comparing the different architectures one can see that \textit{Conv. Dropout} seems to cause higher uncertainties than \textit{Dropout} and \textit{Anch. Dropout} whereas for classification problems, there are only small differences. 
We can see from Fig.~\ref{fig:unc_over_iou} and Fig.~\ref{fig:comp_ep_and_al} that \textit{Aleatoric L2} causes higher uncertainties than \textit{Aleatoric L1} due to modeling the correct likelihood (see Eq.~\ref{eq:loss}).

\subsubsection*{Uncertainty vs. distance}
\readcsv{results/epistemic_unc/dropout_p02_on_baseline_123600/entropy_over_distance.csv}{\dropoutentrodist}
\readcsv{results/epistemic_unc/dropout_p02_on_baseline_123600/tv_over_distance.csv}{\dropouttvodist}
\readcsv{results/epistemic_unc/dropout_p02_on_baseline_123600/unc_over_distance.csv}{\dropoutstddevodist}
\readcsv{results/epistemic_unc/dropout_p02_on_baseline_123600/sincos_over_distance_to_base_angle.csv}{\dropoutbaseangle}
\readcsv{results/epistemic_unc/dropout_p02_conv_second_stage_on_baseline_123600/entropy_over_distance.csv}{\convdropoutentrodist}
\readcsv{results/epistemic_unc/dropout_p02_conv_second_stage_on_baseline_123600/tv_over_distance.csv}{\convdropouttvodist}
\readcsv{results/epistemic_unc/dropout_p02_conv_second_stage_on_baseline_123600/unc_over_distance.csv}{\convdropoutstddevodist}
\readcsv{results/epistemic_unc/dropout_p02_conv_second_stage_on_baseline_123600/sincos_over_distance_to_base_angle.csv}{\convdropoutbaseangle}
\readcsv{results/epistemic_unc/dropout_p02_on_baseline_123600_anchorwise/entropy_over_distance.csv}{\anchdropoutentrodist}
\readcsv{results/epistemic_unc/dropout_p02_on_baseline_123600_anchorwise/tv_over_distance.csv}{\anchdropouttvodist}
\readcsv{results/epistemic_unc/dropout_p02_on_baseline_123600_anchorwise/unc_over_distance.csv}{\anchdropoutstddevodist}
\readcsv{results/epistemic_unc/dropout_p02_on_baseline_123600_anchorwise/sincos_over_distance_to_base_angle.csv}{\anchdropoutbaseangle}
\readcsv{results/kitti_gt_histograms/distance_hist.csv}{\distancegthist}
\readcsv{results/kitti_gt_histograms/distance_to_base_angles_hist.csv}{\distancetobaseanglesgthist}
\readcsv{results/aleatoric_unc/aleatoric_loc_381100/entropy_over_distance.csv}{\locentrodist}
\readcsv{results/aleatoric_unc/aleatoric_loc_381100/tv_over_distance.csv}{\loctvodist}
\readcsv{results/aleatoric_unc/aleatoric_loc_381100/unc_over_distance.csv}{\locstddevodist}
\readcsv{results/aleatoric_unc/aleatoric_loc_381100/sincos_over_distance_to_base_angle.csv}{\locbaseangle}
\readcsv{results/aleatoric_unc/aleatoric_loc_l2_unc_init_185400/entropy_over_distance.csv}{\locltwoentrodist}
\readcsv{results/aleatoric_unc/aleatoric_loc_l2_unc_init_185400/tv_over_distance.csv}{\locltwotvodist}
\readcsv{results/aleatoric_unc/aleatoric_loc_l2_unc_init_185400/unc_over_distance.csv}{\locltwostddevodist}
\readcsv{results/aleatoric_unc/aleatoric_loc_l2_unc_init_185400/sincos_over_distance_to_base_angle.csv}{\locltwobaseangle}

\begin{figure}[ht]
\centering

\begin{tikzpicture}
\tikzstyle{every node}=[font=\small]
\begin{groupplot}[group style={group size=1 by 4, vertical sep=1mm}]

\nextgroupplot[
xmin=0,
xmax=60,
xticklabels={,,},
ytick={0.02, 0.07, 0.12},
yticklabel style={/pgf/number format/fixed, /pgf/number format/precision=2},
ylabel={Shannon entropy},
legend columns=-1,
legend to name=legendepdist,
]
\addplot table [y index=1]{\dropoutentrodist};
\addlegendentry{Dropout Head}; 
\addplot table [y index=1]{\convdropoutentrodist};
\addlegendentry{Conv. Dropout};
\addplot table [y index=1]{\anchdropoutentrodist};
\addlegendentry{Anch. Dropout};

\nextgroupplot[
xmin=0,
xmax=60,
xticklabels={,,},
ymax=0.05,
ytick={0.01, 0.025, 0.04},
yticklabel style={/pgf/number format/fixed, /pgf/number format/precision=3},
scaled y ticks=false,
ylabel={Std. dev. / m},
legend entries={$x$,$y$},
]
\addlegendimage{no markers, gray}
\addlegendimage{no markers, dashed, gray}
\addplot table [y index=1]{\dropoutstddevodist};
\addplot table [y index=1]{\convdropoutstddevodist};
\addplot table [y index=1]{\anchdropoutstddevodist};
\pgfplotsset{cycle list shift=-3};
\addplot+[dashed] table [y index=2]{\dropoutstddevodist};
\addplot+[dashed] table [y index=2]{\convdropoutstddevodist};
\addplot+[dashed] table [y index=2]{\anchdropoutstddevodist};

\nextgroupplot[
xmin=0,
xmax=60,
xticklabels={,,},
ymin=1,
ytick={1.01,1.055, 1.1},
yticklabel style={/pgf/number format/precision=3},
ylabel={Mult. std. dev.},
legend entries={$b_{\mathrm{w}}$,$b_{\mathrm{l}}$},
]
\addlegendimage{no markers, gray}
\addlegendimage{no markers, dashed, gray}
\addplot table [y index=3]{\dropoutstddevodist};
\addplot table [y index=3]{\convdropoutstddevodist};
\addplot table [y index=3]{\anchdropoutstddevodist};
\pgfplotsset{cycle list shift=-3};
\addplot+[dashed] table [y index=4]{\dropoutstddevodist};
\addplot+[dashed] table [y index=4]{\convdropoutstddevodist};
\addplot+[dashed] table [y index=4]{\anchdropoutstddevodist};

\nextgroupplot[
xmin=0,
xmax=60,
xticklabel style={/pgf/number format/precision=0},
ytick={100,450,800},
xlabel={Distance / m},
ylabel={Abs. Frequency},
yticklabel style={/pgf/number format/precision=0},
area style,
]
\addplot+[ybar interval,mark=no,teal!80!black,fill=teal!80!black,fill=teal!80!black,fill=teal!80!black, opacity=0.8] table [y index=1]{\distancegthist};
\end{groupplot}
\end{tikzpicture}
\begin{small}
\ref{legendepdist}
\end{small}
\caption{
Epistemic parameter uncertainties over distance for different configurations in KITTI coordinates.
They correlate with the number of training examples as indicated by the histogram.
}
\label{fig:epistemic_unc_over_dist}
\end{figure}
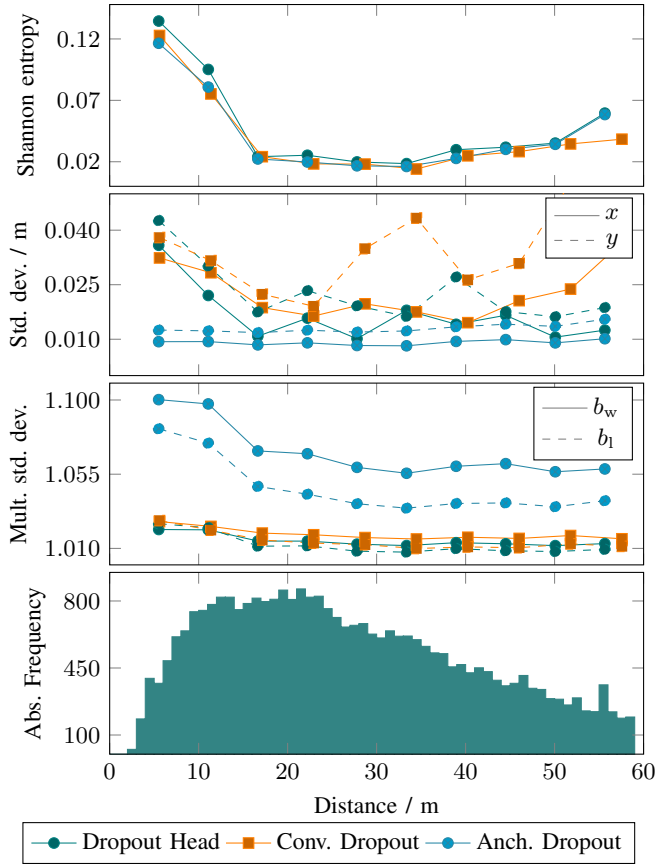

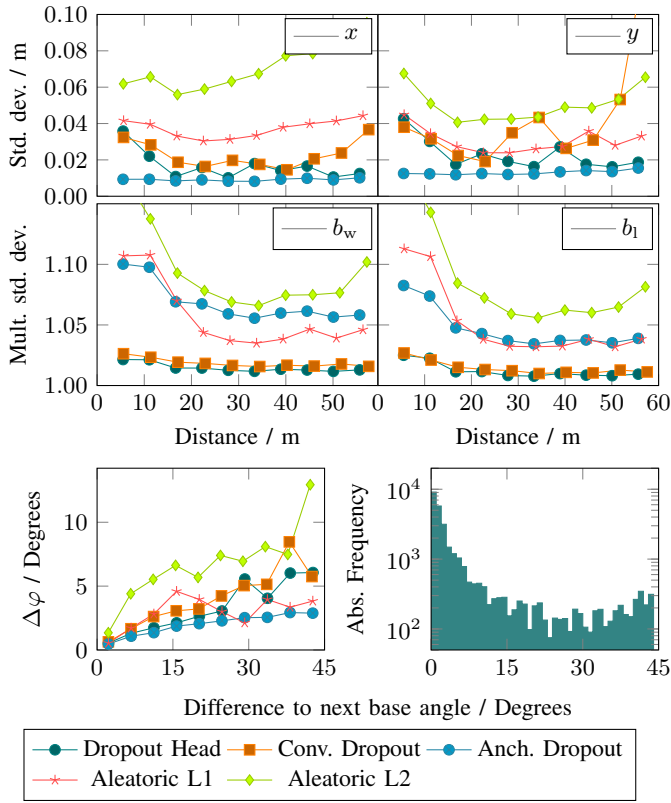
\begin{figure}[ht]
\centering

\begin{tikzpicture}
\pgfplotsset{width=0.6 \columnwidth,}
\tikzstyle{every node}=[font=\small]

\begin{groupplot}[group style={group size=2 by 3, horizontal sep=0mm, vertical sep=1mm}]

\nextgroupplot[
xmax=59.5,
xtick distance=10,
xticklabels={,,},
ymin=0,
ymax=0.1,
yticklabel style={/pgf/number format/fixed, /pgf/number format/precision=2},
ylabel={Std. dev. / m},
legend entries={$x$},
]
\addlegendimage{no markers, gray}
\addplot table [y index=1]{\dropoutstddevodist};
\addplot table [y index=1]{\convdropoutstddevodist}; 
\addplot table [y index=1]{\anchdropoutstddevodist};
\addplot table [y index=1]{\locstddevodist};
\addplot table [y index=1]{\locltwostddevodist}; 

\nextgroupplot[
xmax=60,
xtick distance=10,
xticklabels={,,},
ymin=0,
ymax=0.1,
yticklabels={,,},
legend entries={$y$},
]
\addlegendimage{no markers, gray}
\addplot table [y index=2]{\dropoutstddevodist};
\addplot table [y index=2]{\convdropoutstddevodist}; 
\addplot table [y index=2]{\anchdropoutstddevodist};
\addplot table [y index=2]{\locstddevodist};
\addplot table [y index=2]{\locltwostddevodist};

\nextgroupplot[
xmax=59.5,
xtick distance=10,
xticklabel style={/pgf/number format/precision=0},
xlabel={Distance / m},
ymin=1,
ymax=1.15,
ytick={1,1.05,1.1},
yticklabel style={/pgf/number format/precision=2},
ylabel={Mult. std. dev.},
legend entries={$b_{\mathrm{w}}$},
]
\addlegendimage{no markers, gray}
\addplot table [y index=3]{\dropoutstddevodist};
\addplot table [y index=3]{\convdropoutstddevodist}; 
\addplot table [y index=3]{\anchdropoutstddevodist};
\addplot table [y index=3]{\locstddevodist};
\addplot table [y index=3]{\locltwostddevodist};

\nextgroupplot[
xmax=60,
xtick distance=10,
xticklabel style={/pgf/number format/precision=0},
xlabel={Distance / m},
ymin=1,
ymax=1.15,
ytick={1,1.05,1.1},
yticklabels={,,},
legend entries={$b_{\mathrm{l}}$},
]
\addlegendimage{no markers, gray}
\addplot table [y index=4]{\dropoutstddevodist};
\addplot table [y index=4]{\convdropoutstddevodist};
\addplot table [y index=4]{\anchdropoutstddevodist};
\addplot table [y index=4]{\locstddevodist};
\addplot table [y index=4]{\locltwostddevodist}; 

\nextgroupplot[
yshift=-1cm,
width=0.52\columnwidth,
xshift=-0.04\columnwidth,
xmax=45,
xtick distance=15,
ylabel={$\Delta \varphi$ / Degrees},
xticklabel style={/pgf/number format/fixed, /pgf/number format/precision=0},
yticklabel style={/pgf/number format/fixed, /pgf/number format/precision=0},
legend columns=3,
legend to name=legendbaseangle,
]
\addplot table [y index=3]{\dropoutbaseangle};
\addlegendentry{Dropout Head}; 
\addplot table [y index=3]{\convdropoutbaseangle};
\addlegendentry{Conv. Dropout};
\addplot table [y index=3]{\anchdropoutbaseangle};
\addlegendentry{Anch. Dropout};
\addplot table [y index=3]{\locbaseangle};
\addlegendentry{Aleatoric L1}; 
\addplot table [y index=3]{\locltwobaseangle};
\addlegendentry{Aleatoric L2};

\nextgroupplot[
yshift=-1cm,
width=0.52\columnwidth,
xshift=0.04\columnwidth,
xmax=45,
xtick distance=15,
xticklabel style={/pgf/number format/fixed, /pgf/number format/precision=0},
ylabel={Abs. Frequency},
ymode=log,
ymin=50,
ymax=20000,
area style,
]
\addplot+[ybar interval,mark=no,teal!80!black,fill=teal!80!black,fill=teal!80!black,fill=teal!80!black, opacity=0.8] table [y index=1]{\distancetobaseanglesgthist};
\end{groupplot}
\node (xlabel) at ($(group c1r3.center)!0.5!(group c2r3.center)+(0,-2cm)$) {Difference to next base angle / Degrees};
\end{tikzpicture}

\begin{small}
\ref{legendbaseangle}
\end{small}
\caption{
Comparison of epistemic and aleatoric regression uncertainties (in KITTI coordinates).
Angle uncertainties are calculated for the difference angle to next base angle of $0^{\circ}$ or $90^{\circ}$ and encoded via the maximum differential angle, which results from the uncertainties for $\sin \left( 2 \phi \right)$ and $\cos \left( 2 \phi \right)$.
}
\label{fig:comp_ep_and_al}
\end{figure}

We observe from Fig.~\ref{fig:epistemic_unc_over_dist} that from the sensor origin epistemic uncertainties are very high and strongly decreasing towards 20m.
This may be explained with the small number of training examples close to the sensor and thus poor model confidence. 
For distances >40m epistemic positioning uncertainties increase again due to the decreasing number of training examples.
Compared to the other models \textit{Anch. Dropout} has much lower positioning uncertainties whereas shape uncertainties are much higher.
This results directly from the method chosen to calculate mean and variance. 
Because \textit{Anch. Dropout} computes uncertainties per region proposal, objects from forward runs with inaccurately estimated shapes which would not pass non-maxima-suppression are considered in mean and variance estimation, too. 
On the other hand, object poses always originate from the same proposal explaining small positioning uncertainties.
As stated previously \textit{Conv. Dropout} causes higher uncertainties than \textit{Dropout}.
Fig.~\ref{fig:comp_ep_and_al} compares aleatoric and epistemic regression uncertainties.
Aleatoric uncertainties are much higher than epistemic uncertainties, indicating that the model confidence is higher than sensor precision in most cases.
Furthermore they seem to be more distance-dependent than epistemic uncertainties. 
For distances >20m pose uncertainties increase, which can be explained by the low number of detections per object as also mentioned in \cite{Feng2018}.
For shape uncertainties this effect occurs at greater distances.
For small distances aleatoric uncertainties are high and strongly decreasing. 
We do not think this can be explained with high noise in range sensor data, but also with the small number of training examples for small distances.
Aleatoric uncertainties are learned, too and thus depend on model confidence. 
Epistemic uncertainties in direction of travel ($y$) are higher than in $x$-direction whereas aleatoric uncertainties behave opposite.
This indicates that from range sensor data it is easier to estimate the distance in direction of travel, but the model confidence is smaller because the range of object distances is higher in direction of travel than orthogonal to it.

\subsubsection*{Uncertainty vs. rotation}
In Fig.~\ref{fig:comp_ep_and_al} we depict angular uncertainties over the difference to the next base angle of $0^{\circ}$ or $90^{\circ}$. 
Epistemic uncertainties increase approximately linear with increasing differential angle, indicating that model confidence depends highly on object orientation.
As can bee seen from the histogram this correlates with the number of training examples which are mostly oriented at $0^{\circ}$ or $90^{\circ}$.
Aleatoric uncertainties increase more slowly for larger differential angles.

\subsubsection*{Class uncertainty}
\begin{figure}[ht]
\centering
\pgfplotsset{width=0.6 \columnwidth,}
\begin{tikzpicture}
\tikzstyle{every node}=[font=\small]
\begin{axis}[
hide axis,
xmax=59.5,
xtick distance=10,
ymax=0.1,
legend columns=-1,    
legend to name=legendclasses,
]
\addplot table [y index=1]{\dropoutstddevodist};
\addlegendentry{All Classes}
\addplot table [y index=7]{\dropoutstddevodist}; 
\addlegendentry{Car}
\addplot table [y index=13]{\dropoutstddevodist};
\addlegendentry{Pedestrian}
\addplot table [y index=19]{\dropoutstddevodist};
\addlegendentry{Cyclist}
\end{axis}

\begin{groupplot}[group style={group size=2 by 2, horizontal sep=0mm, vertical sep=1mm}]

\nextgroupplot[
xmax=59.5,
xtick distance=10,
ymax=0.1,
ytick distance=,
legend entries={$x$},
yticklabel style={/pgf/number format/fixed, /pgf/number format/precision=2},
ylabel={Std. dev. / m},
xticklabels={,,},
]
\addlegendimage{no markers, black}
\addplot table [y index=1]{\dropoutstddevodist};
\addplot table [y index=7]{\dropoutstddevodist}; 
\addplot table [y index=13]{\dropoutstddevodist};
\addplot table [y index=19]{\dropoutstddevodist};

\nextgroupplot[
xmax=60,
xtick distance=10,
ymax=0.1,
legend entries={$y$},
xticklabels={,,},
yticklabels={,,},
]
\addlegendimage{no markers, gray}
\addplot table [y index=2]{\dropoutstddevodist};
\addplot table [y index=8]{\dropoutstddevodist}; 
\addplot table [y index=14]{\dropoutstddevodist};
\addplot table [y index=20]{\dropoutstddevodist};

\nextgroupplot[
xmax=59.5,
xtick distance=10,
ymin=1,
ymax=1.09,
xticklabel style={/pgf/number format/precision=0},
yticklabel style={/pgf/number format/precision=2},
legend entries={$b_{\mathrm{w}}$},
xlabel={Distance / m},
ylabel={Mult. std. dev},
]
\addlegendimage{no markers, gray}
\addplot table [y index=3]{\dropoutstddevodist};
\addplot table [y index=9]{\dropoutstddevodist}; 
\addplot table [y index=15]{\dropoutstddevodist};
\addplot table [y index=21]{\dropoutstddevodist};

\nextgroupplot[
xmax=60,
xtick distance=10,
ymin=1,
ymax=1.09,
xticklabel style={/pgf/number format/precision=0},
yticklabel style={/pgf/number format/precision=2},
legend entries={$b_{\mathrm{l}}$},
xlabel={Distance / m},
yticklabels={,,},
]
\addlegendimage{no markers, gray}
\addplot table [y index=4]{\dropoutstddevodist};
\addplot table [y index=10]{\dropoutstddevodist};
\addplot table [y index=16]{\dropoutstddevodist};
\addplot table [y index=22]{\dropoutstddevodist};
\end{groupplot}
\node (title) at ($(group c1r1.center)!0.5!(group c2r1.center)+(0,1.5cm)$) {\textbf{Dropout Head}};
\end{tikzpicture}

\begin{tikzpicture}
\tikzstyle{every node}=[font=\small]
\begin{groupplot}[group style={group size=2 by 2, horizontal sep=0mm, vertical sep=1mm}]

\nextgroupplot[
xmax=59.5,
ymax=0.1,
xtick distance=10,
legend entries={$x$},
yticklabel style={/pgf/number format/fixed, /pgf/number format/precision=2},	
ylabel={Std. dev. / m},
xticklabels={,,},
]
\addlegendimage{no markers, gray}
\addplot table [y index=1]{\locstddevodist};
\addplot table [y index=7]{\locstddevodist}; 
\addplot table [y index=13]{\locstddevodist};
\addplot table [y index=19]{\locstddevodist};

\nextgroupplot[
xmax=60,
ymax=0.1,
xtick distance=10,
legend entries={$y$},
yticklabels={,,},
xticklabels={,,},
]
\addlegendimage{no markers, gray}
\addplot table [y index=2]{\locstddevodist};
\addplot table [y index=8]{\locstddevodist}; 
\addplot table [y index=14]{\locstddevodist};
\addplot table [y index=20]{\locstddevodist};

\nextgroupplot[
xmax=59.5,
ymin=1,
ymax=1.5,
xtick distance=10,
ytick distance=,
xticklabel style={/pgf/number format/precision=0},
yticklabel style={/pgf/number format/precision=1},
legend entries={$b_{\mathrm{w}}$},
xlabel={Distance / m},
ylabel={Mult. std. dev.},
]
\addlegendimage{no markers, gray}
\addplot table [y index=3]{\locstddevodist};
\addplot table [y index=9]{\locstddevodist}; 
\addplot table [y index=15]{\locstddevodist};
\addplot table [y index=21]{\locstddevodist};

\nextgroupplot[
xmax=60,
ymin=1,
ymax=1.4,
xtick distance=10,
ytick distance=,
legend entries={$b_{\mathrm{l}}$},
xlabel={Distance / m},
xticklabel style={/pgf/number format/precision=0},
yticklabel style={/pgf/number format/precision=1},
yticklabels={,,},
]
\addlegendimage{no markers, gray}
\addplot table [y index=4]{\locstddevodist};
\addplot table [y index=10]{\locstddevodist}; 
\addplot table [y index=16]{\locstddevodist};
\addplot table [y index=22]{\locstddevodist};
\end{groupplot}
\node (title) at ($(group c1r1.center)!0.5!(group c2r1.center)+(0,1.5cm)$) {\textbf{Aleatoric L1}};
\end{tikzpicture}
\centering
\ref{legendclasses}
\caption{
Epistemic and aleatoric uncertainties for semantic classes in KITTI coordinates.
Epistemic uncertainties are calculated by \textit{Dropout Head} and aleatoric uncertainties by \textit{Aleatoric L1}.
}
\label{fig:ep_and_al_unc_for_classes}
\end{figure}
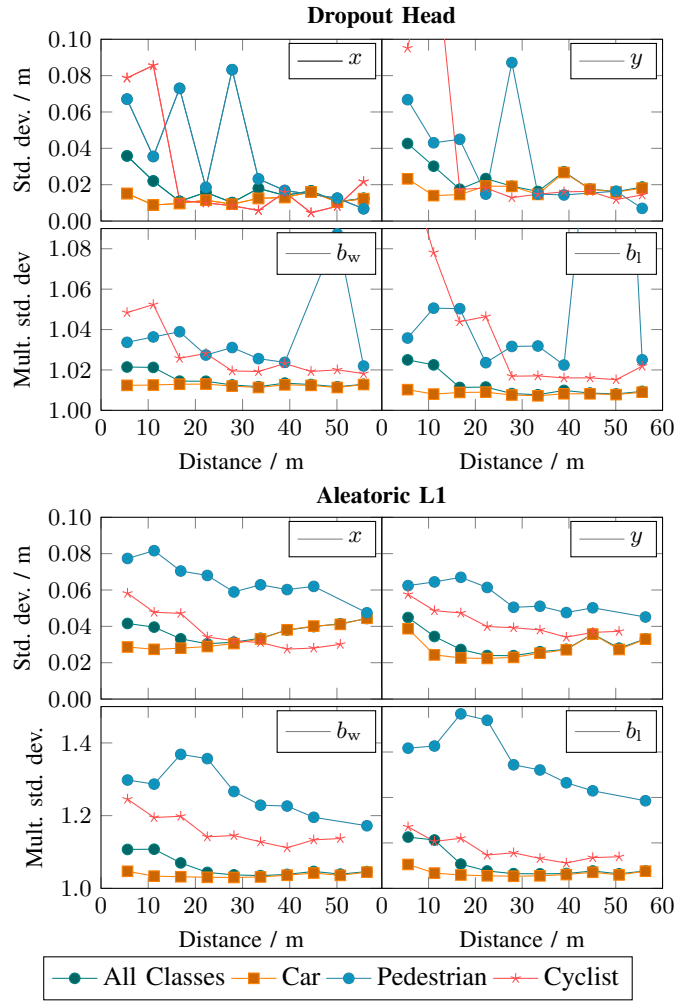

Fig.~\ref{fig:ep_and_al_unc_for_classes} depicts the epistemic and aleatoric uncertainties for all and each of the three classes car, pedestrian and cyclist for \textit{Dropout Head}.
Out of all parameter uncertainties, pedestrians and cyclists are perceived more inaccurate than whereas cars have lower uncertainties.
We suggest this is a direct result of the KITTI training dataset, which has $21,935$ cars but only $3,643$ pedestrians and $1,393$ cyclists.
This could be a reason why the curves in Fig.~\ref{fig:ep_and_al_unc_for_classes} for pedestrians and cyclists are more noisy than for cars, too.
Because overall uncertainty is almost identical to car uncertainty we think that the model primarily learns to detect cars.

Aleatoric uncertainties estimated by \textit{Aleatoric L1} are less noisy but behave qualitatively similar to epistemic uncertainty.
We think this is because small objects like pedestrians have less detections than taller objects such as cars.
This also explains why aleatoric uncertainties for pedestrians are higher than for cyclists.  

\section{Conclusions}
\label{sec:conclusion}

We extended our object detection model to also estimate uncertainties.
By capturing epistemic uncertainty in an efficient way, we gain indicators on the model performance for inputs underrepresented in the training data.
This technique is also helpful for further analysis of data sets.
By estimating aleatoric uncertainty, we are able to explain sensor restrictions which have major impact on the regression parameters.
Therefore, we developed a shape representation which yields to more conservative planning in case of high uncertainty.
As a next step we want to investigate the estimation of uncertainties to all anchors in a single-stage network in order to estimate a probability for unconsidered objects being false-negatives.

\bibliographystyle{IEEEtran}
\bibliography{root}

\end{document}